\documentclass{sigchi}

 \toappear{
}

\usepackage{balance}  
\usepackage{graphics} 
\usepackage[T1]{fontenc}
\usepackage{txfonts}
\usepackage{mathptmx}
\usepackage[pdftex]{hyperref}
\usepackage{color}
\usepackage{booktabs}
\usepackage{textcomp}
\usepackage{microtype} 
\usepackage{ccicons}  

\usepackage{todonotes}

\def\plaintitle{Computational Narrative Intelligence: A Human-Centered Goal for Artificial Intelligence}

\def\emptyauthor{}
\def\plainkeywords{Artificial Intelligence, Machine Learning, Narrative Intelligence, Machine Enculturation}

\makeatletter
\def\url@leostyle{%
  \@ifundefined{selectfont}{
    \def\UrlFont{\sf}
  }{
    \def\UrlFont{\small\bf\ttfamily}
  }}
\makeatother
\def\pprw{8.5in}
\def\pprh{11in}

\definecolor{linkColor}{RGB}{6,125,233}
\hypersetup{%
  pdftitle={\plaintitle},
  pdfauthor={\emptyauthor},
  pdfkeywords={\plainkeywords},
  bookmarksnumbered,
  pdfstartview={FitH},
  colorlinks,
  citecolor=black,
  filecolor=black,
  linkcolor=black,
  urlcolor=linkColor,
  breaklinks=true,
}

\begin{document}

\title{\plaintitle}


\numberofauthors{1}
\author{%
  \alignauthor{%
    \textbf{Mark O. Riedl}\\
    \affaddr{Georgia Institute of Technology} \\
    \affaddr{Atlanta, Georgia, USA} \\
    \affaddr{riedl@cc.gatech.edu} } }

\maketitle

\begin{abstract}
Narrative intelligence is the ability to craft, tell, understand, and respond affectively to stories.
We argue that instilling artificial intelligences with computational narrative intelligence affords a number of applications beneficial to humans.
We lay out some of the machine learning challenges necessary to solve to achieve computational narrative intelligence.
Finally, we argue that computational narrative is a practical step towards machine enculturation, the teaching of sociocultural values to machines.
\end{abstract}


\section{Introduction}

Storytelling is an important part of how we, as humans, communicate, entertain, and teach each other. 
We tell stories dozens of times a day: around the dinner table to share experiences; through fables to teach values; through journalism to communicate important events, and in entertainment movies, novels, and computer games for fun. 
Stories also motivate people to learn, which is why they form the backbone of training scenarios and case studies at school or work.

Despite the importance of storytelling as part of the human experience, computers still cannot reliably create and tell novel stories, nor understand stories told by humans. 
When computers do tell stories, via an eBook or computer game, they simply regurgitate something written by a human. 
They do not partake in the culture we are immersed in, as manifested through journalistic news articles, the movies we watch, or the books we read.

Why does it matter that computers cannot create, tell, or understand stories? 
Artificial intelligence has become more prevalent in our everyday lives. 
Soon, it will not be unusual for us to interact with more advanced forms of {\em Siri} or {\em Cortana} on a daily basis. 
However, when we use those systems today, we find it to be an {\em alien} sort of intelligence. 
The AI makes decisions that sometimes can be hard for us to make sense of. 
Their failures are often due to the fact that they cannot make sense of what we are trying to accomplish or why.

{\em Narrative intelligence} is the ability to craft, tell, understand, and respond affectively to stories.
Research in {\em computational} narrative intelligence seeks to instill narrative intelligence into computers.
In doing so, the goal of developing computational narrative intelligence is to make computers better communicators, educators, entertainers, and more capable of relating to us by genuinely understanding our needs.
Computational narrative intelligence is as much about human-computer interaction as it is about solving hard artificial intelligence problems. 

In this position paper, we enumerate a number of human-centered applications of computational narrative intelligence that may be of benefit to humans interacting with artificial intelligences in the future. 
We also discuss some of the machine learning challenges that will need to be overcome through research to achieve computational narrative intelligence.
Finally, we describe how computational intelligence can provide a way forward to creating artificial intelligences that are more human-like, better at understanding their human users, and more easily comprehended by human users.

\section{Computational Narrative Intelligence}

Winston \cite{winston11} argues that narrative intelligence is one of the abilities that sets humans apart from other animals and non-human-like artificial intelligences.
Research in computational narrative intelligence has sought to create computational intelligences that can answer questions about stories, generate fictional stories and news articles, respond affectively to stories, and represent the knowledge contained in natural language narratives.

Given that humans communicate regularly and naturally though narratives, one of the long-standing challenges of computational narrative intelligence has been to answer questions about stories \cite{schank77,mueller04,weston15}.
Question-answering is a way of verifying that a computer is able to understand what a human is saying.
However, question-answering about narrative content is considered to be more challenging than fact-based question-answering due to the causal and temporal relationships between events, which can be complex and are often left implicit. 
One prerequisite for narrative question-answering is a better understanding of how to represent the knowledge contained in natural language narratives \cite{chambers08,finlayson11,elson12}.

The flip-side of understanding stories is the creation of novel, fictional story content such as fairy tales and computer game plots \cite{gervas09,riedl:jair2010,porteous10,zhu14}.
The obvious application of fictional story generation is entertainment.
On-demand narrative generation can maintain a continuous flow of novel content for users to engage with while customizing the content to individual preferences and demands.
One may imagine serial novels, serial scripts for TV shows and movies, or serial quests and plotlines in computer games.
However, note that even entertainment can convey morals and other pedagogical aspects.

Computational narrative intelligences can also create plausible sounding---but fictional---stories that might happen in the real world \cite{swanson12,li:aaai2013}.
While not meant to be entertaining, the generation of plausible real world stories provides a strong, objective measure of general computational intelligence. 
Plausible real-world story generation can be used to generate virtually unlimited scenarios for skill mastery in training simulations \cite{zook:fdg2012}.
Computational narrative intelligences could engage in forensic investigations by hypothesizing about sequences of events that have not been directly observed.
Virtual agents, such as virtual health coaches, can appear more life-like and create rapport with humans by sharing fictional vignettes and gossip \cite{bickmore-iva09}.

Computational narrative intelligence also brings computers one step closer to understanding the human experience and predicting how humans will respond to narrative content.
Automated journalists generate narrative texts about real world events and data such as sports and financial reports (e.g., \cite{statsmonkey}). 
Automated journalists may benefit from narrative intelligence when determining how best to convey a narrative to different audiences.
Going beyond journalism, it is important to note that humans can have very visceral emotional responses to stories.
Understanding how the human will interpret and respond to narrative situations has important implications if we wish for computers to avoid accidentally making people upset or anxious.
Computers may one day intentionally attempt to induce pleasure, or create a sense of suspense \cite{oneill:aaai2014} in both entertainment and journalistic contexts.


Finally, narrative can be used to explain the behavior of artificial intelligences. 
Any process or procedure can be told as a narrative, so it follows that an AI can describe the means by which it came to a conclusion or the reasons why it performed an action by couching its explanation in narrative terms.
As part of a explanatory process, narratives can convey counterfactuals---what would have happened if circumstances had been different.
We hypothesize that narrative explanation will be more easily understood by non-expert human operators of artificial intelligence since the human mind is tuned for narrative understanding.

 \section{Machine Learning Challenges}
 
Automated story understanding and automated story generation have a long history of pursuit in the field of artificial intelligence. 
Until recently, most approaches used hand-authored formal models of the story world domain the generator or understander would operate in \cite{schank77,meehan77,lebowitz87,mueller04,gervas05,riedl:jair2010}.
This made {\em open-domain} narrative intelligence---the sort employed by humans---intractable due to knowledge engineering bottlenecks.
More recent approaches use machine learning to attempt to automatically acquire and reuse domain models from narrative corpora on the Internet \cite{McIntyre10,chambers08,swanson12} and from crowdsourcing \cite{li:aaai2013}.

There are at least four primary challenges related to learning domain models from narrative corpora and using them to create stories or explanations.
First, human-written narratives are written to be consumed by other humans.
We use {\em theory of mind} to infer what others are likely to already know and adjust our storytelling accordingly.
Thus, human-written narratives collected into a corpus often leave out elements that are assumed to be commonly shared knowledge among other humans but possibly not known by computers.
For example, a news corpus may have a story about bank robbery, but that story only has the points that make it unique from other bank robberies and ``newsworthy.'' 
In some sense, all stories interesting enough to tell, or to have been told, are outliers from each other, making patterns hard to detect.
A machine learning system would never learn about the aspects of the domain model that are most common to all bank robberies.

Stories are often told to highlight an unexpected obstacle or event in an otherwise typical situation.
By virtue of telling a story of this sort, one may infer the counterfactual as the norm \cite{schubert11}.
Crowdsourcing allows for greater control of the narrative content and can be used to acquire a corpus of {\em typical} stories about situations at the desired level of granularity \cite{li:aaai2013}.
Many children's books and television shows teach expectations for common situations such as going to a doctor's office or what to expect on the first day of school. 

Second, narrative intelligence is closely associated with commonsense reasoning.
It is necessary for both narrative understanding and for narrative generation.
Humans learn commonsense knowledge and reasoning through a lifetime of experiences in the real world.
Learning commonsense knowledge as been an ongoing challenge in AI and machine learning.

Commonsense knowledge in the form of declarative facts and procedures will be essential in comprehending narratives.
Research into automated commonsense knowledge acquisition includes \cite{lenat95,liu04,etzioni11,mitchell15}.
Images and video also implicitly capture commonsense knowledge (e.g., things fall downward, people kick balls but not bricks, etc.) \cite{parikh15} and techniques that jointly learn from stories with accompanying video or illustrations may provide key insights.

Third, natural language stories written by humans for humans make abundant use of metaphors and metonymy~\cite{lakoff87}.
Decoding the meaning of metaphors and metonymy requires high-level semantic comprehension of the narratives collected into a machine learning corpus.

A few research projects have attempted to use metaphor in the automated generation of stories \cite{hervas07,veale14}.
Hobbs~\cite{hobbs92} lays out three general approaches to understanding metaphors: transferring properties from one entity to another, mapping aspects of one thing to another by inference, or mapping aspects of one thing to another by analogy.
Analogical mapping has received the most attention in computational narrative intelligence \cite{falkenhainer89,riedl:mm2010,zhu14}.

Fourth, creating stories requires a model of {\em creativity} as process that transcend straightforward pattern learning.
The space of all possible, tellable, and interesting stories is vast.
This is one explanation for why the generation of stories via sampling from recurrent neural networks trained on narrative corpora has not fared well to date because of the complexity of human-written narratives and the need for very large training sets.
Further, stories make use of long-term causal connections between events that have not been easy to model; long term dependencies mean that stories, and the process of creating stories, are non-Markovian.
However, some recent progress has been made in using long short-term memory neural nets that can extract script-like representations from text \cite{pichotta15}. 

Descriptions of human creativity emphasize the blending of two or more mental models to create new concepts.
The appeal of {\em conceptual blending} \cite{fauconnier98} is the invention of concepts that might never have existed in a data set or even the real world.
Conceptual blending shares similarities to unsupervised transfer learning, a critical area of research in machine learning.
One example in the domain of creativity is the blending of two neural nets trained on different aspects of art \cite{stylenet}.
However, an equivalent approach has not yet been found for story generation.

Solving these challenges will be necessary in order to achieve a complete, open-domain, computational narrative intelligence that is trained from narrative corpora.
In some cases, the challenges are those associated with semantic-level natural language processes.
However note that television, movies, dramatic plays, comic books, and illustrated children's books can also be sources of valuable data and require integrated natural language processing and machine vision. 

\section{Machine Enculturation}
 
In addition to the applications described ealier---and assuming the above challenges can be met---computational narrative intelligence may present a way forward toward {\em machine enculturation} \cite{riedl:aaai-ethics2016}.
Machine enculturation is the act of instilling social norms, customs, values, and etiquette into computers so that they can (a) more readily relate to us and (b) avoid harming us (physically or psychologically) or creating social disruptions.
In a perfect world, humanity would come with a user manual that we could simply scan into a computer. 
However, for any sufficiently complex domain, such as the real world, manually encoding a comprehensive set of values or rewards in order to recreate sociocultural behavior is intractable.

If sociocultural values are not easily instilled in artificial intelligences, perhaps they can be learned.
Instead of a user manual we have the collected works of fiction by different cultures and societies. 
This collected works give us examples with which to teach an artificial intelligence the ``rules'' of our societies and cultures. 
These stories includes the fables or allegorical tales passed down from generation to generation, such as the tale of George Washington confessing to chopping down a cherry tree. 
Fictional stories meant to entertain can be viewed as examples of protagonists existing within and enacting the values of the culture to which they belong, from the mundane---eating at a restaurant---to the extreme---saving the world. 

Stories are an effective means of conveying complex tacit and experiential knowledge that implicitly encodes social and cultural values \cite{bruner91}.
Humans do not need to be trained to communicate via storytelling, nor be trained to decode the knowledge contained within narratives.
Computers will likely require require human-level narrative comprehension to mine social and cultural values from fictional and non-fictional narrative texts because those values are rarely made explicit.

The actions of characters in stories can be viewed as demonstrations of socioculturally appropriate behavior under hypothetical situations.
Unlike demonstrations, which occur in the environment that the artificial intelligence will operate in, narratives may be more general.
This presents some new challenges. 
Stories written in natural language can contain events and actions that are not executable by an artificial intelligence. 
Stories are written by humans for humans and thus make use of commonly shared knowledge, leaving many things unstated. 
Stories frequently skip over events that do not directly impact the telling of the story, and sometimes also employ flashbacks, flashforwards, and achrony which may confuse an artificial learner.
However, learning from narratives can make certain things easier. 
Stories can make explicit the normally unobservable mental operations and thought processes of characters.
Written stories make dialogue more explicit in terms of whom is speaking, although some ambiguity remains \cite{elson10} and comprehension of language is still be an open challenge. 

Machine enculturation may give us a way forward toward achieving artificial intelligences that understand humans better, and make themselves more comprehensible---less alien---to humans.
Further, agents and robotics that act in accordance with social values will naturally avoid situations where humans will be harmed or inconvenienced whenever possible.
Harrison and Riedl \cite{riedl:aaai-ethics2016} describe a technique, {\em learning from stories} (LfS), for emulating human behavior expressed in simple, crowdsourced narratives.
It is proof of concept that machine enculturation may be feasible via machine learning over a corpus of stories. 
  
\section{Conclusions} 

Narrative intelligence is central to many of the things we as humans do, from communication to entertainment to learning.
Narrative is also an effective means of storing and disseminating culture.
In this position paper we argue that future artificial intelligences should be instilled with computational narrative intelligence so that they can act like humans, or understand human wants, needs, and desires.
Artificial intelligences instilled with computational narrative intelligence may be more effective at communicating with humans and explaining their behavior.
Finally, computational narrative intelligence may be a practical step towards machine enculturation.

\balance{} 

\bibliographystyle{SIGCHI-Reference-Format}
\bibliography{hcc,riedl}

\end{document}